\pdfoutput=1
\documentclass[11pt]{article}

\usepackage[final]{coling}

\usepackage{times}
\usepackage{latexsym}

\usepackage[T1]{fontenc}
\usepackage[utf8]{inputenc}

\usepackage{microtype}

\usepackage{inconsolata}

\usepackage{graphicx}

\usepackage{amssymb}
\usepackage{amsmath}

\usepackage{algorithm}
\usepackage{algpseudocode}
\usepackage{listings}
\usepackage{xcolor}
\usepackage{xurl}

\title{Stop Jostling: Adaptive Negative Sampling Reduces the Marginalization of Low-Resource Language Tokens by Cross-Entropy Loss%
\thanks{Accepted at LoResLM 2025, the First Workshop on Language Models for Low-Resource Languages, co-located with COLING 2025.}}

\author{Galim Turumtaev \\
  \texttt{turumtaev.gz@gmail.com} \\
}

\begin{document}
\renewcommand{\labelenumii}{\arabic{enumi}.\arabic{enumii}}
\maketitle
\begin{abstract}
Neural language models often struggle with low-resource languages due to the limited availability of training data, making tokens from these languages rare in the training set. This paper addresses a specific challenge during training: rare tokens are disproportionately affected by marginalization, which prevents them from learning effectively. We propose a thresholding technique that reduces the impact of this marginalization, allowing rare tokens to benefit from more meaningful alignment. Through experiments with a character-level language model, we demonstrate that this method significantly improves performance on low-resource language validation data. This work is the first to show how negative sampling can be applied to improve the representation of rare tokens by limiting the harmful influence of excessive marginalization, offering a new approach to enhancing language model performance for underrepresented languages.
\end{abstract}

\section{Introduction}

\begin{figure}[t]
  \includegraphics[width=\columnwidth]{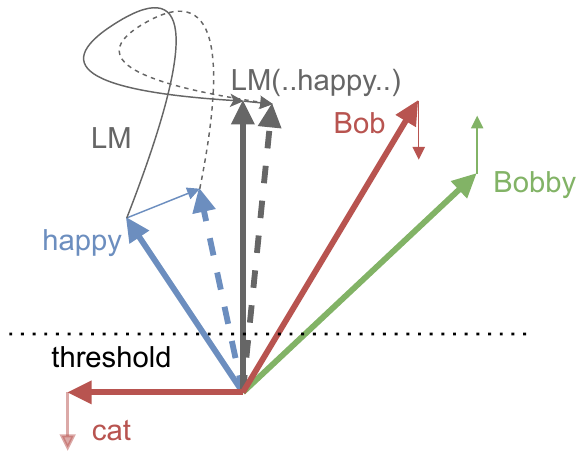}
  \caption{Three embeddings optimization types. Example for $X$ = \textit{'don't worry, be happy' by}; $Y$~=~\textit{Bobby McFerrin}. \textbf{Context alignment (blue)}: adjust $w_{\text{happy}}$ so that $g_{\theta}(\dots, w_{\text{happy}}, \dots)$ moves closer to $w_{\text{Bobby}}$. \textbf{Target alignment (green)}: move $w_{\text{Bobby}}$ closer to $g_{\theta}(\dots, w_{\text{happy}}, \dots)$. \textbf{Non-target marginalization (red)}: move non-relevant $w_{\text{Bob}}$ and $w_{\text{cat}}$ away from $g_{\theta}(\dots, w_{\text{happy}}, \dots)$. The proposed method prevents marginalization of embeddings under the threshold.}
  \label{fig:example-scenarios}
\end{figure}

Neural language models have revolutionized natural language processing (NLP), providing state-of-the-art results in a wide range of tasks, such as machine translation, text generation, and sentiment analysis. However, the effectiveness of these models heavily relies on the availability of large, high-quality datasets for pre-training. This dependency presents a significant challenge for low-resource languages, which often lack the extensive corpora needed for effective language model training.

One of the main issues faced by multilingual language models is the difficulty in learning effective representations for tokens from low-resource languages. These tokens, which occur infrequently during training, tend to receive less alignment and are more responsive to noise from irrelevant contexts. Recent studies have highlighted how this imbalance can negatively impact model performance \cite{chang2024when}. Existing solutions often focus on improving the general quality of embeddings \cite{gao2018representation} or limiting the influence of rare tokens on the overall training process \cite{yu-etal-2022-rare}.

In this paper, we identify a specific source of noise that affects rare tokens, which we call \textbf{marginalization}. Marginalization by cross-entropy loss pushes non-target embeddings away from irrelevant contexts and disproportionately impacts rare tokens, preventing them from learning meaningful representations. Unlike previous methods that address the impact of rare tokens on the overall model, we focus on reducing the negative impact \textit{on} rare tokens themselves, which is a less explored but equally important problem.

To address this issue, we propose a simple yet effective adaptive negative sampling technique, which we call \textbf{thresholding}. By applying a threshold to the logits, the model effectively ignores non-relevant tokens during the training process, allowing those non-relevant tokens to receive more meaningful updates. This approach is novel in its use of negative sampling to improve the representations of unselected negative samples, rather than focusing solely on training efficiency or contrastive learning.

We validate thresholding effectiveness through experiments with a character-level multilingual language model trained on simulated mixed low-resource and high-resource language data. The results demonstrate that the proposed technique improves the representation and performance of rare tokens, making it particularly valuable for enhancing language models in low-resource settings.

The main contributions of this paper are as follows:

\begin{itemize}
    \item Identification of \textbf{marginalization} as a key factor degrading the quality of rare token representations (§\ref{sec:Problem}).
    \item Introduction of a \textbf{thresholding} technique to mitigate marginalization (§\ref{sec:Proposition}), with experiments showing improvements in language model performance on low-resource data (§\ref{sec:Experiments}).
\end{itemize}

By addressing the challenges faced by tokens in low-resource languages, this paper presents a novel approach to improving multilingual language model performance, contributing to more balanced progress in NLP for underrepresented languages.

\section{Problem}
\label{sec:Problem}

\subsection{Intuition}

Let us begin with an example. Consider the task of language modeling where the input prompt is the title of a song: \textit{'Don't Worry Be Happy' by}. The goal is to predict the next word.

The correct continuation is \textit{Bobby}, completing the sentence \textit{'Don’t Worry Be Happy' by Bobby McFerrin.} Now, let us reflect on the learning process of a language model:

\begin{enumerate}
    \item \textbf{Was anything new learned about the word \textit{Bobby}?} Yes, it was learned that \textit{Bobby} is the nickname of the artist who performed \textit{Don't Worry Be Happy}.
    \item \textbf{Was anything new learned about the words \textit{don’t}, \textit{worry}, \textit{be}, \textit{happy} and \textit{by}?} Yes, it was learned that this sequence of words may be followed by \textit{Bobby} in this context.
    \item \textbf{Was anything new learned about the word \textit{Bob}?} This song is often incorrectly attributed to Bob Marley. Now it was learned that \textit{Bob} is not the correct nickname here.
    \item \textbf{Was anything new learned about the word \textit{cat}?} No, there is no new information about cats in this context.
\end{enumerate}

In summary, while the model learns valuable associations between the correct words, irrelevant words, such as \textit{cat}, should not be influenced by this example. Yet, because of \textbf{cross-entropy} loss, many modern language models still “learn” representations of irrelevant words, even when they don’t belong in the context.

This issue relates to the \textbf{distributional hypothesis}, which states that words occurring in similar contexts tend to have similar meanings. Ideally, the model should learn word representations based only on relevant context. However, when using \textbf{cross-entropy} as the loss function, modern models tend to "push" non-target words, such as \textit{cat}, slightly away from the model’s last hidden state. Although this "push" is often small, in the case of rare tokens or low-resource languages, it can degrade the learned representations.

\subsection{Formalization}

Consider a vocabulary \( V = \{v_1, v_2, \dots, v_N\} \) of size \( N \), and an embedding matrix \( W = [w_1, w_2, \dots, w_N] \), where row \( w_i \) corresponds to token \( v_i \) for each \( i \in \{1, \dots, N\}\). A training sentence is denoted as \( (x_0, x_1, \dots, x_M) \), with length \( M+1 \), where \( x_i \in V \) for each \( i \in \{0, \dots, M\}\). The last hidden state before the classification head, \( h_t \), is produced by the model's body \( g_\theta \) with parameters \( \theta \), based on the first \( t \) input tokens:

\[
h_t = g_\theta(w_{x_0}, \dots, w_{x_{t-1}})
\]

When using \textbf{weight tying} \cite{press-wolf-2017-using}, the probability of the token \( x_t \) is calculated by the language model as:

\[
P_{\theta}(x_t|x_0, \dots, x_{t-1}) = \frac{\exp(\langle h_t, w_{x_t} \rangle)}{\sum_{i=1}^N \exp(\langle h_t, w_i \rangle)}
\]

Where \(\langle a, b \rangle\) denotes the dot product of vectors \(a\) and \(b\). During training, the cross-entropy loss:

\[
\mathcal{L}_\theta(x_t) = -\log \left( P_\theta(x_t|x_0, \dots, x_{t-1}) \right)
\]

is minimized for all \( t \in \{1, \dots, M\}\).

Embeddings \( W \) are optimized simultaneously in three distinct ways:

\begin{enumerate}
    \item \textbf{Context alignment}: For all \( k \in \{0, \dots, t-1\} \), \( w_{x_k} \) is optimized to maximize \( \frac{\exp(\langle g_\theta(\dots, w_{x_k}, \dots), w_{x_t} \rangle)}{\sum_{i=1}^N \exp(\langle g_\theta(\dots, w_{x_k}, \dots), w_i \rangle)}\). The gradient is \(\frac{\partial \mathcal{L}_\theta(x_t)}{\partial h_t}\frac{\partial h_t}{\partial w_{x_k}} \).
    \item \textbf{Target alignment}: \( w_{x_t} \) is optimized to maximize \( \langle h_t, w_{x_t} \rangle \). The gradient is \(\frac{\partial \mathcal{L}_\theta(x_t)}{\partial w_{x_t}} \).
    \item \textbf{Non-target marginalization}:  For all \( v_i \in V \), where \( v_i \neq x_t \), \( w_i \) is optimized to minimize \( \langle h_t, w_i \rangle \). The gradient is \(\frac{\partial \mathcal{L}_\theta(x_t)}{\partial w_i} \).
\end{enumerate}

In the example in Figure~\ref{fig:example-scenarios}, various tokens, including irrelevant ones such as \textit{cat}, are affected by this third type of optimization, which we refer to as \textbf{marginalization}. As we show in §\ref{sec:influence_of_marginalization}, this noise may be significant for rare tokens and tokens from low-resource languages.

\section{Method}
\label{sec:Proposition}
\subsection{Algorithm}

To reduce marginalization, we propose a \textbf{thresholding} technique that is applied to the logits after the language model's classification head but before calculating the cross-entropy loss.

Let us revisit the song example. Assume that the model assigns probabilities as follows: \( P_\theta(\textit{Bob}) \gtrsim P_\theta(\textit{Bobby}) \gg P_\theta(\textit{cat}) \). Although it makes sense to lower the probability of \textit{Bob}, the probability of \textit{cat} is already very low and can be ignored. This allows the embedding of \textit{cat} to align better in its own relevant contexts.

The core idea is to ignore the tokens \( v_i \) with \( P_\theta(v_i) \ll P_\theta(x_t) \). This is achieved by thresholding logits based on a selected \texttt{margin} as described in Algorithm~\ref{alg:threshold_logits}.

\begin{algorithm}[h]
\caption{Thresholding Logits}
\label{alg:threshold_logits}
\begin{algorithmic}[1]
\State \textbf{Input:} $logits, x, margin$
\For{each $t$ in $[1, \dots, M]$}
    \State $threshold_t \gets (logits_t[x_t] - margin)$
    \For{each $i$ in $[1, \dots, N]$}
        \If{$logits_t[v_i] < threshold_t$}
            \State $logits_t[v_i] \gets -\infty$
        \EndIf
    \EndFor
\EndFor
\end{algorithmic}
\end{algorithm}

A simple and effective implementation of this algorithm in the PyTorch framework can be found in Appendix~\ref{sec:appendix-torch}.

By applying this thresholding, the probabilities \( P_{\theta}(v_i) < P_{\theta}(x_t) \times e^{-\texttt{margin}} \) are effectively set to 0. This makes the marginalization gradients zero for the corresponding embeddings.

Referring back to Figure~\ref{fig:example-scenarios}, after applying the threshold, the logits for the token \textit{cat} become \( -\infty \), so it is no longer marginalized. However, the logits for another token, \textit{Bob}, remain above the threshold, meaning \textit{Bob} will still be marginalized.

\subsection{Hyperparameter}
\label{sec:Hyperparameter}

This method introduces a new hyperparameter \texttt{margin}. Although we do not cover the optimal choice of \texttt{margin} in this work, we provide an idea on how to limit the search range for \texttt{margin}.

Although the \texttt{margin} theoretically can be set to any value between 0 and \( +\infty \), it is clear that as \(\texttt{margin}~\to~+\infty\), the proposed method converges to standard \textbf{cross-entropy} loss. A large \texttt{margin} will have little to no effect on the performance of the model.

On the other hand, as \(\texttt{margin}~\to~0\), there will be a long tail of irrelevant tokens with small \( P_{\theta} \), that was not marginalized enough. Due to their large number, they will noticeably reduce \( P_{\theta}(x_t) \), increasing the model's perplexity. We describe this phenomenon in more detail in §\ref{sec:Experiments}.

Between these two extremes, there may be a range of suitable \texttt{margin} values that improve the representation of rare tokens without significantly affecting performance on frequent tokens.

Let \( P_{\theta, T}(v_i) \) represent the probability of the token \( v_i \) after applying the temperature \( T \). In Appendix~\ref{sec:appendix}, we show that by choosing 
\[
\texttt{margin}~=~T\times \ln\left( \frac{(N-1) \times \texttt{top\_p}}{1 - \texttt{top\_p}} \right)
\]

thresholding will not affect the tokens \( v_i \) with \( P_{\theta, T}(v_i) \) within the \texttt{top\_p} distribution of \( P_{\theta, T} \). 

Given the widespread use of nucleus sampling in modern language models, this may be sufficient to offset the negative impact on frequent tokens while still benefiting rare tokens. For example, for nucleus sampling with \( T=0.9, \texttt{top\_p} = 0.99 \), and vocabulary size \( N = 100,000 \), choosing 
\begin{align*}
\texttt{margin} &= 0.9 \times \ln\left( \frac{(100,000-1) \times 0.99}{1 - 0.99} \right)\\
                & \approx 14.50 
\end{align*}
ensures that tokens appearing in the top 0.99 of the \( P_{\theta, T} \) distribution will always be marginalized.

This choice may still be too conservative and might not provide enough improvement for rare tokens. However, this relatively low upper bound should make it easier to find a balanced \texttt{margin} between 0 and \( T \times \ln\left( \frac{(N-1) \times \texttt{top\_p}}{1 - \texttt{top\_p}} \right) \).

Similarly, for min-\(p\) sampling \cite{nguyen2024turningheatminpsampling}, by choosing \(\texttt{margin} = T \times \ln\left(p_{\text{base}}\right)\), thresholding will not affect the tokens \(v_i\) within the \(\mathcal{V}_{\text{min}}\) set of min-\(p\) sampling, where \(p_{\text{base}}\) is a hyperparameter of min-\(p\) sampling.

\section{Experiments}
\label{sec:Experiments}

The code is available on GitHub\footnote{\url{https://github.com/turumtaev/StopJostling}}. For this project, we used the \texttt{nanoGPT} implementation by Karpathy\footnote{\url{https://github.com/karpathy/nanoGPT}}. We conducted experiments on a small dataset of Shakespeare's texts and trained a character-level language model. The dataset contains 65 unique characters, with a highly imbalanced distribution (Figure~\ref{fig:token-distribution}).

\subsection{Data and Model}

\begin{figure}[t]
  \includegraphics[width=\columnwidth]{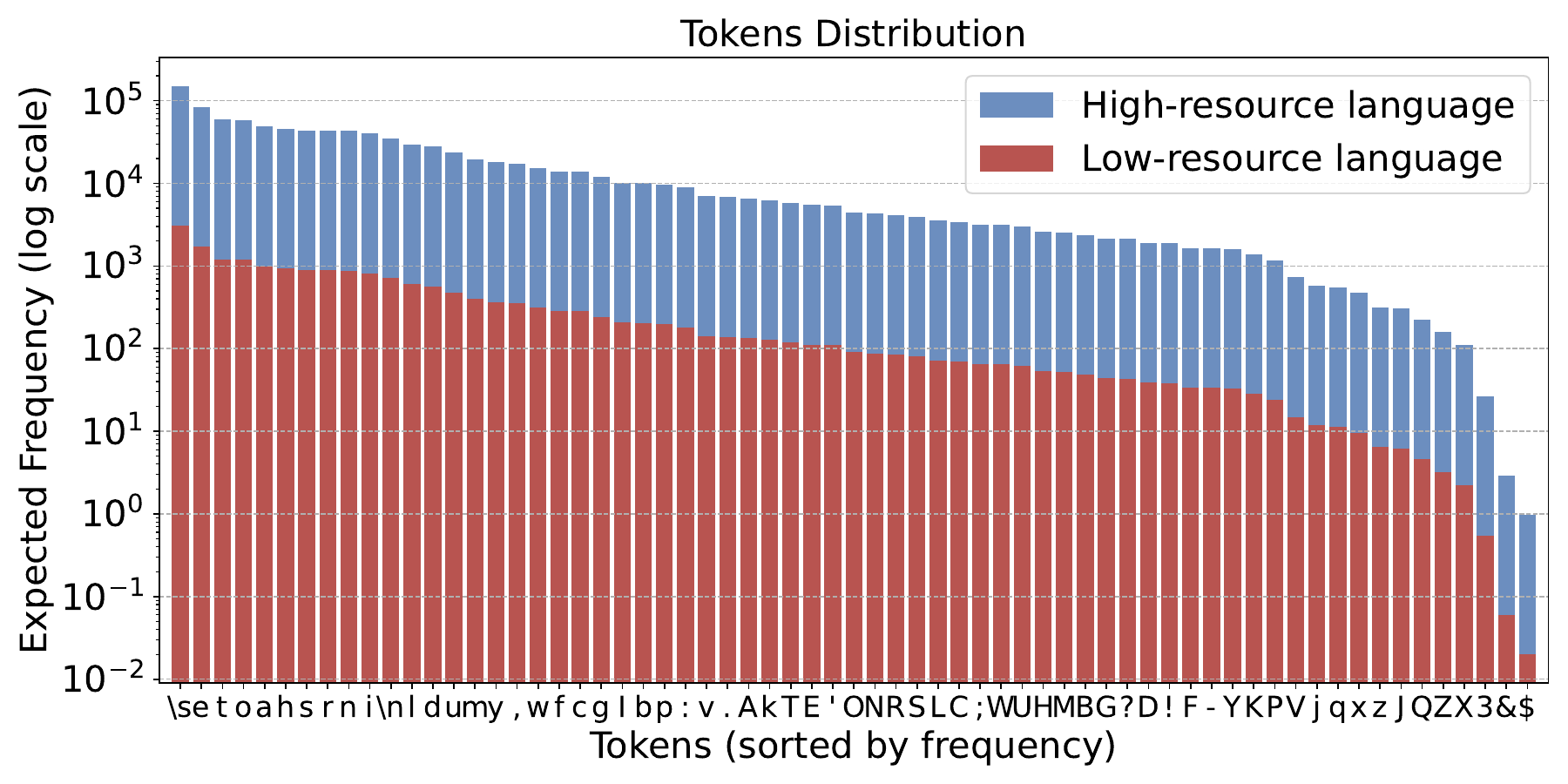}
  \caption{Token distribution in the experiment with a simulated low-resource (2\%) and high-resource (98\%) languages. For the high-resource language, expected frequencies range from 150,209.5 (token " ") to 0.98 (token "\$"); for the low-resource language, expected frequencies range from 3,065.5 to 0.02.}
  \label{fig:token-distribution}
\end{figure}

The Shakespeare dataset provides a toy example with significant imbalance in token occurrences. To simulate both high- and low-resource languages, following \cite{K2020Cross-Lingual}, we modified the character-level tokenizer. In 2\% of randomly selected training sentences, we added \(N=65\) to character IDs, simulating a second "low-resource" language with token IDs ranging from 65 to 129. We selected this 2\% ratio for the low-resource language as it is small enough to observe the negative impact of marginalization, yet realistic, as the second most popular language in GPT-3 pre-training data (French) accounts for about 2\% of words\footnote{\url{https://github.com/openai/gpt-3/blob/master/dataset_statistics/languages_by_word_count.csv}}. Figure~\ref{fig:token-distribution} shows the token distribution for both high-resource and low-resource simulations.

The following models were evaluated:
\begin{itemize}
    \item \textbf{Baseline:} A GPT-2 architecture model with 800k parameters and weight tying.
    \item \textbf{Monolingual:} The baseline model trained solely on low-resource language data (2\% of the training steps).
    \item \textbf{Proposed:} The baseline model with thresholding applied, tested with margins between 0 and 8 (approximating $1 \times \ln\left(\frac{(N-1) \times 0.95}{1 - 0.95}\right)$).
    \item \textbf{Proposed+SE:} The proposed model with separated embeddings for better handling by the AdamW optimizer (more details in §\ref{sec:Separated-embeddings}).
\end{itemize}

The exact hyperparameters used in the experiment are listed in Table~\ref{tab:hyperparameters} in the Appendix.

\subsection{Influence of Marginalization}
\label{sec:influence_of_marginalization}

\begin{figure}[t]
  \includegraphics[width=\columnwidth]{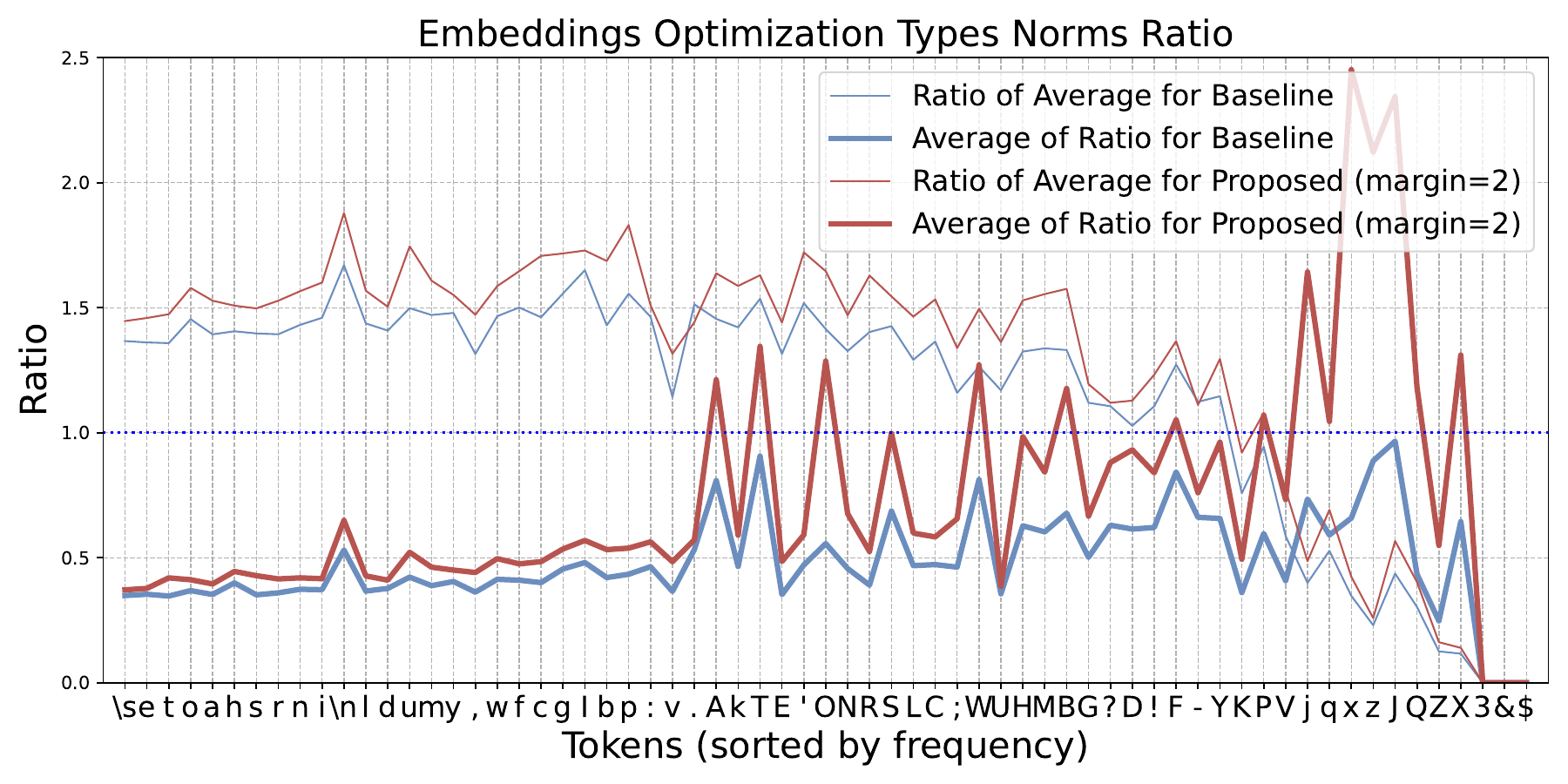}
  \caption{Ratio of embedding gradients norms for different tokens in the low-resource language. Tokens are sorted by frequency. Rare tokens have a lower \textbf{Ratio of Average}. In the baseline model, all tokens have an \textbf{Average of Ratio} below 1, indicating that \textbf{marginalization} has a strong effect on these tokens. The proposed method increases the \textbf{Average of Ratio} by 45\% and the \textbf{Ratio of Average} by 12\% on average.}
  \label{fig:optimization-ratio}
\end{figure}

First, we measured the influence of marginalization on each token. For all tokens, we interpreted the token's embedding optimization as the sum of the 3 types of optimization described above. For each embedding, we calculated its gradient from backpropagation as the sum of gradients from the 3 types of optimization. Knowing the gradients for each token and optimization type, we logged \( \lVert grad_{type,i,step} \rVert \) — the norms of the gradients with type \(type\) for embedding \( w_i \) in step \(step\). Then, we calculated the following ratios:

\begin{itemize}
    \item \textbf{Ratio of Average}:
    \[
    \frac{\text{avg}_{s}(\lVert grad_{1,i,s} \rVert + \lVert grad_{2,i,s} \rVert)}{\text{avg}_{s}(\lVert grad_{3,i,s} \rVert)}
    \]
    \item \textbf{Average of Ratio}:
    \[
    \text{avg}_{s}\left( \frac{\lVert grad_{1,i,s} \rVert + \lVert grad_{2,i,s} \rVert}{\lVert grad_{3,i,s} \rVert} \right)
    \]
\end{itemize}

Both ratios for tokens from the low-resource language are plotted in Figure~\ref{fig:optimization-ratio}, with tokens sorted by frequency. It is clear that the problem of marginalization exists for low-resource language data: the 14 least frequent tokens have \textbf{Ratio of Average} below 1, and all tokens have an \textbf{Average of Ratio} below 1. This indicates that for these tokens, the influence of marginalization is significant. The proposed method increases both ratios\footnote{The proposed method makes \(\lVert grad_{3,i,s} \rVert = 0\) for some \((v_i, s)\), making it impossible to calculate the \textbf{Average of Ratio}. For such \((v_i, s)\), \(\lVert grad_{3,i,s} \rVert\) was estimated with \((\text{avg}_s(\lVert grad_{3,i,s} \rVert)\), which provides a lower bound for the \textbf{Average of Ratio}}.

\subsection{Results and Observations}
\begin{table*}[!ht]
  \centering
  \begin{tabular}{|l|c|c|c|c|c|c|c|}
    \hline
    \textbf{Model} & \textbf{Lang.} & \textbf{PPL} & $\textbf{PPL}_{best}(\text{T}_{best})$ & \textbf{Accuracy} & \textbf{Recall@5} & \textbf{MRR} & \textbf{I(W)} \\
    \hline
    \textbf{Baseline} & HR & 5.04 & 5.01 (1.08) & 0.5187 & 0.8299 & 0.6541 & 0.8422 \\
    & LR & 10.65 & 10.63 (0.95)  & 0.3147 & 0.6883 & 0.4803 & 0.4173 \\
     \hline
    \textbf{Monolingual} & HR & - & - & - & - & - & - \\
    (LR data only) & LR & 12.24 & 12.08 (0.89) & 0.2851 & 0.6632 & 0.4543 & \textbf{0.8917} \\
    \hline
    \textbf{Proposed} & HR & 5.02 & 5.00 (1.07) & 0.5212 & 0.8296 & 0.6557 & 0.8394 \\
    (\texttt{margin}=8) & LR & 10.69 & 10.67 (0.95) & 0.3127 & 0.6907 & 0.4801 & 0.4450 \\
    \hline
    \textbf{Proposed} & HR & \textbf{5.01} & \textbf{4.99} (0.94) & 0.5231 & 0.8313 & 0.6571 & 0.8381 \\
    (\texttt{margin}=4) & LR & 10.90 & 10.67 (0.87)  & 0.3244 & 0.6882 & 0.4871 & 0.6479 \\
    \hline
    \textbf{Proposed} & HR & 5.82 & 5.07 (0.71) & 0.5291 & 0.8336 & 0.6614 & 0.8499\\
    (\texttt{margin}=2) & LR & 11.56 & 9.62 (0.67)  & 0.3581 & 0.7166 & 0.5161 & 0.6422 \\
    \hline
    \textbf{Proposed} & HR & 9.33 & 5.13 (0.48) & \textbf{0.5297} & 0.8344 & \textbf{0.6626} & 0.8884\\
    (\texttt{margin}=1) & LR & 17.30  & 9.54 (0.46) & 0.3714 & 0.7204 & 0.5255 & 0.7651 \\
    \hline
    \textbf{Proposed} & HR & 5.24 & 5.24 (\textbf{1.02}) & 0.5241 & 0.8323 & 0.6579 & 0.8207 \\
    (\texttt{margin}=1, T=0.46) & LR & 10.46  & 10.46 (\textbf{1.00}) & 0.3459 & 0.7064 & 0.5060 & 0.6730 \\
    \hline
    \textbf{Proposed+SE} & HR & 5.86 & 5.08 (0.71) & 0.5254 & 0.8318 & 0.6595 & 0.8892 \\
    (\texttt{margin}=2) & LR & \textbf{10.41}  & 8.41 (0.65) & 0.3947 & 0.7417 & 0.5478 & 0.7092 \\
    \hline
    \textbf{Proposed+SE} & HR & 9.18 & 5.12 (0.48) & 0.5274 & 0.8357 & 0.6615 & 0.8706 \\
    (\texttt{margin}=1) & LR & 13.91  & 6.90 (0.44) & 0.4544 & 0.7827 & 0.5986 & 0.7109 \\
    \hline
    \textbf{Proposed+SE} & HR & 14.94 & 5.15 (0.34) & 0.5273 & \textbf{0.8364} & 0.6614 & \textbf{0.8929} \\
    (\texttt{margin}=0.6) & LR & 19.94  & \textbf{6.17} (0.32) & \textbf{0.4868} & \textbf{0.8090} & \textbf{0.6277} & 0.7619 \\
    \hline
  \end{tabular}
  \caption{Evaluation metrics for models on the validation dataset for high-resource (HR) and low-resource (LR) languages. The best result for each metric and language is bolded. As expected, a \texttt{margin} of 8 is too conservative and has minimal impact on performance. Metrics improve as the \texttt{margin} decreases, achieving the best result with a carefully selected \(\texttt{margin}=0.6\). Applying temperature scaling \(T=T_{best}\) during training helps improve the perplexity \textbf{PPL}. The use of \textbf{Separated Embeddings (SE)} shows a significant improvement in model performance on low-resource languages.}
  \label{tab:Metrics}
\end{table*}

The models were compared on the validation data using the following metrics:
\begin{itemize}
    \item \textbf{PPL}: Character-level perplexity of \(P_{\theta}(x_t)\).
    \item \(\textbf{PPL}_{best} (T_{best})\): Best \textbf{PPL} of \(P_{\theta, T}(x_t)\) among different \(T\). The proposed method increases \(P_{\theta}\) for the unreliable tail of tokens, and applying a lower temperature \(T\) typically helps. See §\ref{sec:Long-tail-problem} for more details and Appendix~\ref{sec:ppl-temperature} for the proposed method to reduce the problem.
    \item \textbf{Accuracy, Recall@5, and Mean Reciprocal Rank (MRR)}: Although Baseline outperforms thresholded models in terms of \textbf{PPL}, there is other evidence suggesting that this is mainly due to an unreliable tail. The thresholded models rank the target token higher, even for high-resource language tokens.
    \item \textbf{I(W)}: Following \cite{mu2018allbutthetop}, anisotropic embeddings may harm the quality of the language model \cite{yu-etal-2022-rare}. Thresholded models show better \textbf{I(W)} values, providing more isotropic embeddings.
\end{itemize}

Table~\ref{tab:Metrics} shows the results. Starting with a safe \texttt{margin} of 8, we observe an improvement in quality for low-resource languages as the margin decreases. The proposed method suffers from an unreliable tail, but training \(P_{\theta,1/T_{best}}(v_i)\) may help reduce the problem, with slightly worse results for other metrics (Appendix~\ref{sec:ppl-temperature}). \textbf{Separated Embeddings (SE)} (§\ref{sec:Separated-embeddings}) further improve the performance of the language model in low-resource languages.

\subsection{Long tail of tokens}
\label{sec:Long-tail-problem}

\begin{figure}[t]
  \includegraphics[width=\columnwidth]{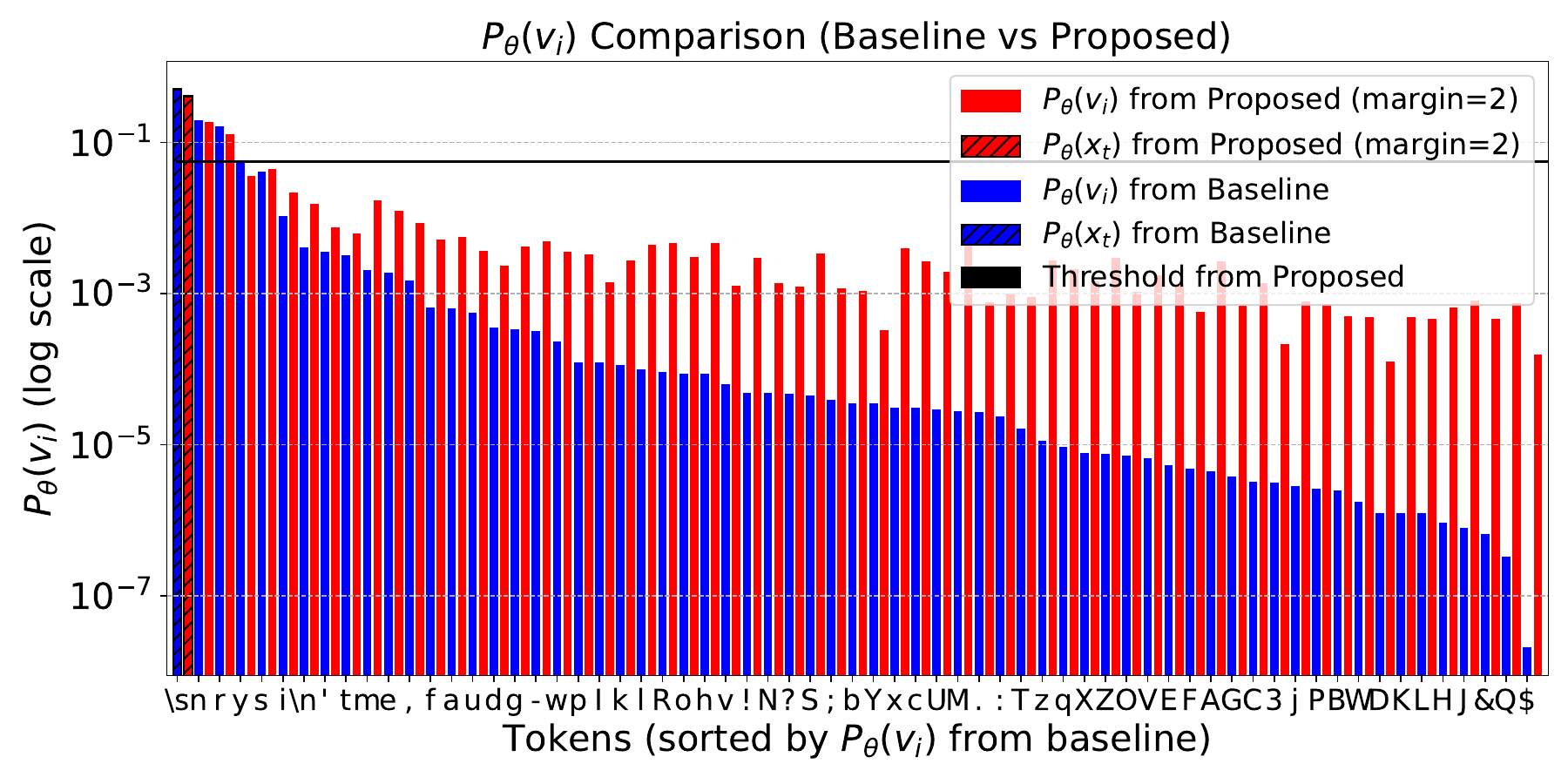}
  \caption{Example of real distribution of $P_{\theta}(v_i)$ from the Baseline and Proposed methods. Due to thresholding,  $P_{\theta}(v_i)$ for non-relevant tokens is pushed down only until they fall below the threshold. This creates the issue of an unreliable tail, where even though $P_{\theta}(x_t)$ from the Proposed remains the highest among all tokens, its value is still lower than that of $P_{\theta}(x_t)$ from the Baseline.}
  \label{fig:long_tail_example}
\end{figure}

\textbf{PPL} is a widely used metric to evaluate language models. However, thresholding naturally increases \textbf{PPL} due to the presence of a long tail of tokens. Figure~\ref{fig:long_tail_example} illustrates how the long tail of non-relevant tokens with higher probabilities can reduce the probability of the target token, thereby increasing the \textbf{PPL}. The thresholding process marginalizes these non-relevant tokens only until their probabilities fall below the threshold, creating what we refer to as an unreliable tail.

Today sampling methods such as nucleus sampling are widely used. Such methods exclude the long tail of tokens from generation. Similarly, reducing the temperature helps suppress the probability of the tail. In our experiments, we observe that after applying the optimal temperature \(T_{best}\), the proposed method achieves a lower \(\textbf{PPL}_{best}\) compared to the baseline.

\subsection{Separated Embeddings}
\label{sec:Separated-embeddings}

In Figure~\ref{fig:optimization-ratio}, we observe the ratio of embedding gradient norms. However, in practice, the actual ratio differs due to the use of the \texttt{AdamW} optimizer and its momentum calculations, which average gradients over multiple steps. Even after applying thresholding, the gradients applied are never exactly zero: \texttt{AdamW} treats embeddings as rows of a single matrix, meaning that if at least one embedding has non-zero gradients, the momentum for all embeddings will be updated, and those updated momenta will be applied.

To avoid this issue, we modified the setup by saving the embeddings as a list of weights, with each token having its own independent embedding vector. This allows \texttt{AdamW} to skip updates for an embedding \(w_i\) if there are no new gradients specifically for token \(v_i\). This approach effectively isolates the updates for each token, ensuring that the optimization only affects tokens with relevant gradients.

Experiments show that using separated embeddings significantly improves model quality, with improvements in several key metrics: $\times 0.72$ \(\textbf{PPL}_{best}\), $\times 1.22$ \textbf{Accuracy}, $\times 1.09$ \textbf{Recall@5}, and $\times 1.14$ \textbf{MRR}.

It should be noted that, unlike thresholding, \textbf{SE} only affects the optimization of \textbf{unselected} tokens. The significant improvement in quality by \textbf{SE} suggests that this improvement occurs precisely by reducing marginalization, and not by contrastive learning.

A simple implementation of the Separated Embeddings layer in the PyTorch framework is provided in Appendix~\ref{sec:appendix-torch}.

\subsection{Learned "Translations"}
\begin{figure*}[t]
  \includegraphics[width=\linewidth]{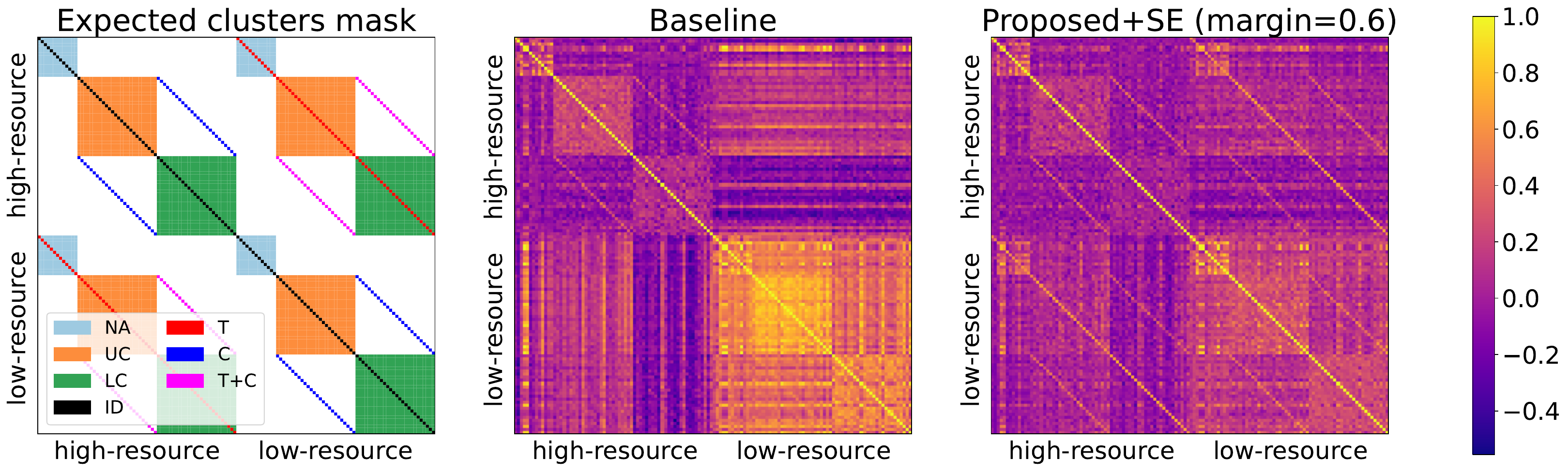}
  \caption{\textbf{Expected clusters mask and cosine similarity of embeddings}. The mask highlights clustering patterns: \textbf{NA} (13 non-alphabetical characters), \textbf{UC} (26 uppercase letters), \textbf{LC} (26 lowercase letters), \textbf{ID} (identity diagonal, always 1), \textbf{T} (translation of the same letter across languages), \textbf{C} (capitalization of the same letter), and \textbf{T+C} (capitalization of the same letter across languages). The embeddings are sorted by language and then alphabetically. The baseline model tends to marginalize low-resource (LR) embeddings, pushing them in the same direction. It only learns clusters and capitalization patterns for the high-resource language. In contrast, the proposed model captures all relationships described by the mask, revealing meaningful connections between characters across languages, \textbf{without any parallel corpus} in training data.}
  \label{fig:cosine_similarity}
\end{figure*}

\begin{figure}
  \includegraphics[width=\columnwidth]{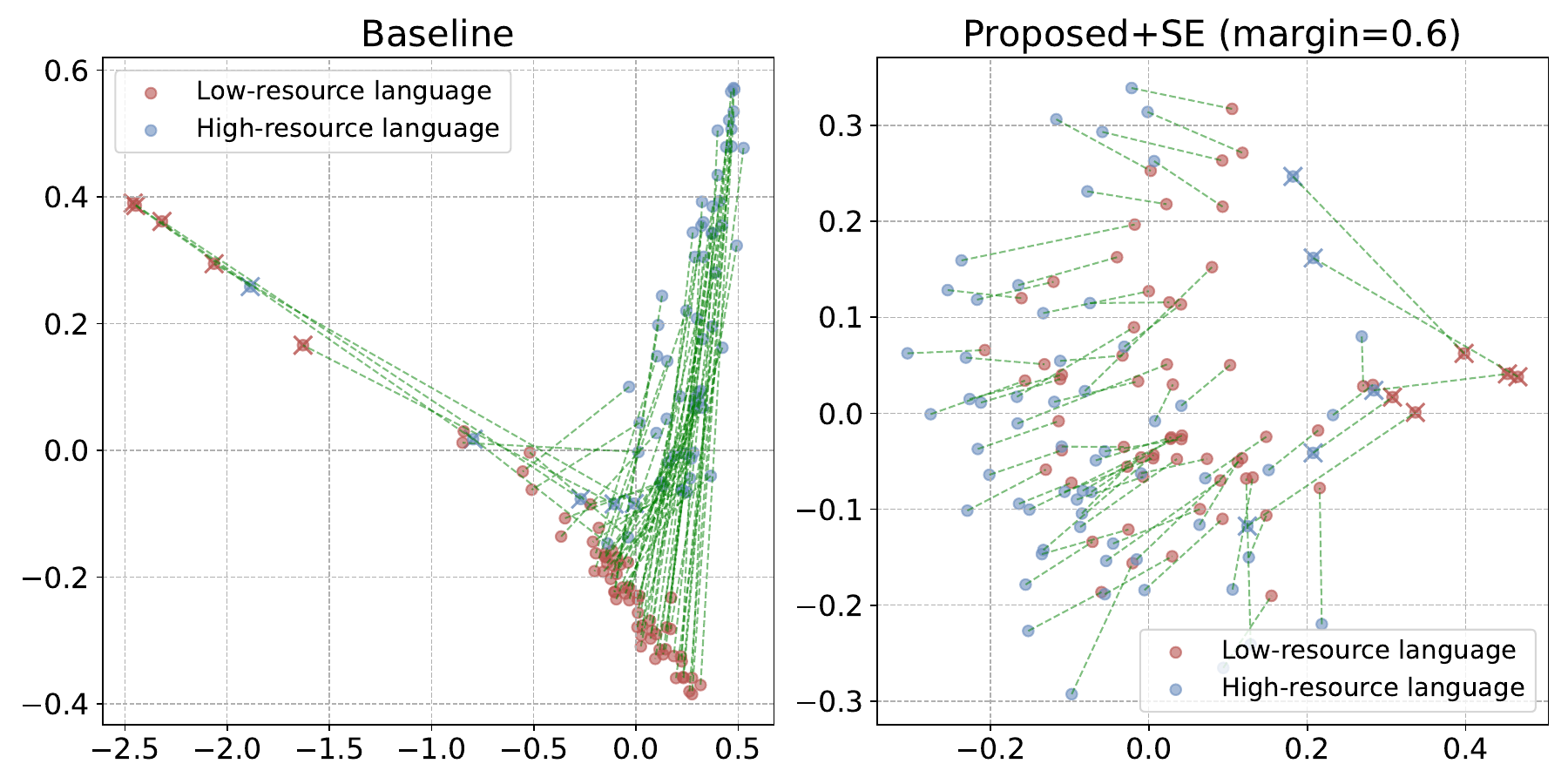}
  \caption{Comparison of PCA decomposition of embeddings. Embeddings from different languages are differentiated by color. The five embeddings of the rarest characters of each language are marked with crosses. Embeddings of the same character from 2 languages are connected with green lines.}
  \label{fig:PCA_embeddings}
\end{figure}

\begin{table*}[!ht]
  \centering
  \begin{tabular}{|l|c|c|c|c|}
    \hline
    \textbf{Model} & \(\textbf{A}_{HR}\) & \(\textbf{a}_{HR}\) & \(\textbf{A}_{LR}\) & \(\textbf{a}_{LR}\)\\
    \hline
    \textbf{Baseline} & \(\text{B}_{HR}\) (0.41) & \(\text{o}_{HR}\) (0.38) & \(\text{e}_{LR}\) (0.82) & \(\text{i}_{LR}\) (0.86) \\
     & \(\text{E}_{HR}\) (0.37) & \(\text{i}_{HR}\) (0.35) & \(\text{O}_{LR}\) (0.81) & \(\text{o}_{LR}\) (0.71) \\
     & \(\text{O}_{HR}\) (0.36) & \(\text{e}_{HR}\) (0.34) & \(\text{I}_{LR}\) (0.77) & \(\text{e}_{HR}\) (0.71) \\
     \hline
   \textbf{Proposed+SE} & \(\textbf{A}_{LR}\) (0.54) &  \(\textbf{a}_{LR}\) (0.69) & \(\textbf{A}_{HR}\) (0.54) & \(\textbf{a}_{HR}\) (0.69) \\
    (\texttt{margin}=0.6) & \(\textbf{a}_{HR}\) (0.36) &  \(\textbf{A}_{HR}\) (0.36) & \(\text{B}_{LR}\) (0.45) & \(\text{i}_{LR}\) (0.44) \\
     & \(\textbf{a}_{LR}\) (0.32) &  \(\textbf{A}_{LR}\) (0.28) & \(\textbf{a}_{LR}\) (0.43) & \(\textbf{A}_{LR}\) (0.43) \\
     \hline
  \end{tabular}
  \caption{Top-3 neighbors by cosine similarity for embeddings of characters from high-resource (HR) and low-resource (LR) languages. In this example, the proposed method places capitalizations and "translations" among the top-3 neighbors in 10 out of 12 cases.}
  \label{tab:translations}
\end{table*}

Figures~\ref{fig:cosine_similarity} and~\ref{fig:PCA_embeddings} demonstrate that the proposed method helps the model to learn meaningful relationships between characters in different languages. It brings the embeddings of the same character from high-resource and low-resource languages closer together.

Table~\ref{tab:translations} presents the top-3 neighbors based on cosine similarity for the embeddings of characters \textit{A} and \textit{a} from both high-resource (HR) and low-resource (LR) languages. The model with thresholding and SE places capitalizations and "translations" of the character as the top-3 neighbors in 10 out of 12 cases, whereas the baseline model does so in 0 out of 12 cases.

This evidence shows that the proposed method not only improves ranking metrics but also helps the model learn more meaningful character representations across languages without any parallel corpus in the training data.

\section{Related work}

\citet{chang2024when} show that the addition of too much multilingual data can negatively impact the performance of language models in low- and high-resource languages due to limited model capacity, a phenomenon known as the \textbf{curse of multilinguality}.

\citet{gao2018representation} explore the \textbf{representation degeneration problem}, where token embeddings degenerate into a narrow cone, reducing the capacity of the model. To address this, they proposed \textbf{cosine regularization} to increase the expressiveness of the embeddings. Similarly, \citet{zhang-etal-2020-revisiting} propose using \textbf{laplacian regularization} to tackle the same issue.

\citet{yu-etal-2022-rare} link the \textbf{representation degeneration problem} with \textbf{anisotropy} and use \textbf{I(W)} to measure \textbf{isotropy}. The authors identify specific parts of the negative log likelihood loss gradient as the main cause of the problem, which aligns with the ideas presented in this paper. In addition, they propose \textbf{adaptive gradient gating (AGG)}. While the concept of \textbf{AGG} is similar to the thresholding technique proposed in this paper, \textbf{AGG} is more complex and requires counting token frequencies during training.

\textbf{Negative sampling (NS)} is widely used in many machine learning tasks \cite{10454000}. \textbf{NS} helps reduce computational complexity in tasks with large or "infinite" sample spaces, such as images or word-level tokenizers. For example, \citet{mikolov2013efficientestimationwordrepresentations} introduced random sampling to select negative samples during Word2Vec training. \textbf{NS} is also commonly used in contrastive learning: \citet{godey2024headless} proposed \textbf{contrastive weight tying (CWT)}, which uses in-batch tokens as negatives. Contrastive learning is widely used to train sentence-level embeddings (\citealp{feng-etal-2022-language}; \citealp{wang2024textembeddingsweaklysupervisedcontrastive}; \citealp{sturua2024jinaembeddingsv3multilingualembeddingstask}). \citet{wang2024bridging} show how contrastive learning and in-batch negative sampling help to reduce the "language gap".

We applied some of the methods proposed in these related works and compared them with our approach. The corresponding metrics and comparisons are provided in the Appendix.

\section{Conclusion}

In this paper, we propose a method to improve the performance of language models in low-resource languages by reducing the impact of marginalization through logit thresholding.

The experimental results demonstrate significant improvements. The language modeling accuracy for the low-resource language increased from 0.31 with baseline to 0.49, which is close to the accuracy for the high-resource language (0.53). Additionally, the $\textbf{PPL}_{best}$ for the low-resource language was reduced from 10.63 to 6.11, almost reaching the $\textbf{PPL}_{best}$ for the high-resource language (4.97). The proposed approach not only improves performance metrics but also helps the model learn better representations, as evidenced by the alignment of "translations" of the same characters across different languages.

Furthermore, while previous work on negative sampling has primarily focused on enhancing training efficiency or improving the representation of positive examples, this method is, to the best of our knowledge, the first to show how negative sampling can directly improve the representation of non-sampled tokens.

\section{Limitations of the work}

We conducted experiments only with a small model and dataset. This introduces several limitations to the work.

\textbf{Data and model size:} Experiments with larger models could potentially alter the results. \citet{chang2024when} show that increasing the size of the model tends to improve the performance on multilingual data. At the same time, in Appendix~\ref{sec:appendix-margin-and-d-model}, we share our intuition about why a larger embedding dimension size could enhance the positive effect of thresholding.

\textbf{Tokenizer:} Using different tokenization techniques, such as Byte Pair Encoding (BPE), could affect the outcome. Since BPE alters the token distribution, \citet{zouhar-etal-2023-tokenization} demonstrate that the performance of the model correlates with the entropy of the token distribution generated by the tokenizer.

\textbf{Languages:} We tested the proposed method only on simulated multilingual data. Testing with real languages might lead to different results. \citet{chang2024when} also show that adding data from similar languages improves model performance more than adding data from dissimilar languages. In our simulated setup, each character in the original language has a corresponding character in the simulated language with exactly the same meaning. This 1:1 correspondence does not exist in natural multilingual data. However, we are optimistic that our method will still perform well with natural languages. As shown in Table 2, our thresholding approach brings lowercase and uppercase forms of the same character closer together. Importantly, capitalization does not rely on a 1:1 mapping. Based on this evidence, we believe thresholding has potential for success in real-world multilingual scenarios.

\textbf{Downstream performance:} While the proposed method shows a significant improvement in the validation data, this does not necessarily guarantee improved performance in downstream tasks. Further testing on various downstream tasks is needed to confirm the method's effectiveness.

\textbf{Model architecture:} Although in our experiments we use a decoder transformer architecture, the method is not restricted to it. Since it modifies logits, a common component in many architectures, this method could also be applied to other model types.

\textbf{Weight tying:} While our explanation is tailored to models with weight tying, the method is not limited to such models. The results of a similar model without weight tying can be found in Appendix~\ref{sec:weight-tying}.

\textbf{Comparison with other methods:} An extended comparison of metrics for further modifications of the proposed method, along with comparisons to related work, is available in Table~\ref{tab:modifications-quality} in the Appendix. However, not all methods from related works were implemented or hyperparameter-tuned well.

\bibliography{custom}

\appendix

\section{Estimation of \texttt{margin}}
\label{sec:appendix}

Modern language models commonly utilize sampling techniques such as top-\(k\) sampling or nucleus sampling to eliminate the "unreliable tail" of low-probability tokens \cite{Holtzman2020The}. By setting a \texttt{top\_p} and temperature \( T \) for nucleus sampling, with a sufficient \texttt{margin}, the model can be trained so that \( P_{\theta}(v_i) \) is optimized until \( P_{\theta, T}(v_i) \) falls outside of \texttt{top\_p}. Here, \( P_{\theta, T}(v_i) \) represents the probability after applying temperature \( T \).

\textbf{Lemma 1}: 
If there exists at least one token $v_k$, such that $P_{\theta}(v_k) > P_{\theta}(v_i)$ and
\[
P_{\theta}(v_i) < \frac{1 - \texttt{top\_p}}{N-1},
\]
then \( v_i \) is outside of \texttt{top\_p} in nucleus sampling.

\textbf{Proof}: 
Assume the contrary — that \( v_i \) is inside \texttt{top\_p}. This implies that \( v_i \) is not among the lowest probability tokens that make up \( 1 - \texttt{top\_p} \) of the distribution. Consequently, there exist \( n \) other tokens \( \{v_{j_1}, \dots, v_{j_n}\} \) such that for any \( v_j \), \( P_{\theta}(v_j) \leq P_{\theta}(v_i) \), and
\[
1 - \texttt{top\_p} < P_{\theta}(v_i) + \sum_j P_{\theta}(v_j).
\]
Since \( v_i \) and \( v_k \) cannot be \( v_j \), it follows that \(n \leq N-2\) and
\begin{align*}
1 - \texttt{top\_p} &< P_{\theta}(v_i) + \sum_j P_{\theta}(v_j) \\
                    &\leq (N - 1) \times P_{\theta}(v_i) \\
                    &< (N - 1) \times \frac{1 - \texttt{top\_p}}{N-1}.
\end{align*}

This results in a contradiction; therefore, \( v_i \) must be outside of \texttt{top\_p}.

\textbf{Lemma 2}:
If token \( v_i \) has been thresholded, then:
\[
P_{\theta, T}(v_i) < P_{\theta, T}(x_t) \, e^{-\texttt{margin}/T}.
\]

\textbf{Proof}:
\begin{align*}
P_{\theta, T}(v_i) &= \frac{P_{\theta}(v_i)^{1/T}}{\sum_{V} P_{\theta}(v_k)^{1/T}} \\
                   &< \frac{\left( P_{\theta}(x_t) \, e^{-\texttt{margin}} \right)^{1/T}}{\sum_{V} P_{\theta}(v_k)^{1/T}} \\
                   &= P_{\theta, T}(x_t) \, e^{-\texttt{margin}/T}.
\end{align*}

\textbf{Lemma 3}

For a \texttt{margin} defined as \(\texttt{margin} = T\times\ln\left( \frac{(N-1) \times \texttt{top\_p}}{1 - \texttt{top\_p}} \right)\),
any thresholded token \( v_i \) will have \( P_{\theta, T}(v_i) \) outside of the \texttt{top\_p} distribution.

\textbf{Proof:}

- \textbf{Case 1}: If \( P_{\theta, T}(x_t) \geq \texttt{top\_p} \), then \( P_{\theta, T}(v_i) \) cannot be in \texttt{top\_p}, as \( P_{\theta, T}(x_t) \) already accounts for at least \texttt{top\_p} of the probability mass.

- \textbf{Case 2}: If \( P_{\theta, T}(x_t) < \texttt{top\_p} \), then, according to Lemma 2:
\begin{align*}
P_{\theta, T}(v_i) &< P_{\theta, T}(x_t) \, e^{-\texttt{margin}/T} \\
                   &< \texttt{top\_p} \, e^{ - T \ln\left( \frac{(N - 1) \times \texttt{top\_p}}{1 - \texttt{top\_p}} \right)/T } \\
                   &= \texttt{top\_p} \, \frac{(1 - \texttt{top\_p})}{(N - 1) \times \texttt{top\_p}} \\
                   &= \frac{1 - \texttt{top\_p}}{N-1}.
\end{align*}

Since \( P_{\theta, T}(v_i) < \frac{1 - \texttt{top\_p}}{N-1} \) and \(P_{\theta, T}(v_i) < P_{\theta, T}(x_t)\), according to Lemma 1, \( P_{\theta, T}(v_i) \) must be outside of \texttt{top\_p}.

This \texttt{margin} estimation allows us to limit the search range between 0 and \(T ~\times~\ln\left( \frac{(N-1) \times \texttt{top\_p}}{1 - \texttt{top\_p}} \right)\), which increases slowly as the vocabulary size \( N \) increases.

\section{\texttt{margin} and \texttt{d\_model}}
\label{sec:appendix-margin-and-d-model}

\begin{table*}[!ht]
  \centering
  \begin{tabular}{|l|c|c|c|c|c|}
    \hline
    \textbf{Model} & \texttt{d\_model} & \(\texttt{margin}_\text{init}(\sigma)\) & \(\texttt{margin}_\text{PT}(\sigma)\) & \(\alpha_\text{init}(\sigma)\) & \(\alpha_\text{PT}(\sigma)\) \\
    \hline
    Small & 768 & 14.04 (0.02) & 32.96 (5.4) & 0.940 (1.6e-3) & 0.029 (4.5e-3) \\
    Medium & 1024 &  18.93 (0.02) & 49.05 (8.66) &  0.948 (1.2e-3) & 0.043 (7.1e-3) \\
    Large & 1280 & 23.83 (0.02) & 70.04 (2.75) & 0.954 (9.5e-4) & 0.816 (0.047) \\
    XL & 1600 & 29.99 (0.02) & 75.52 (2.28) & 0.958 (7.7e-4) & 0.840 (0.041) \\
    \hline
  \end{tabular}
  \caption{ Estimation for effective \texttt{margin} and \(\alpha\) for pre-trained (PT) and randomly initialized models from GPT-2 family. Increasing \texttt{d\_model} increases the estimation of effective \texttt{margin} from Appendix~\ref{sec:appendix-margin-and-d-model}. For initialized models, \(\alpha\) doesn't change much with the increase of \texttt{d\_model}; after pre-training, \(\alpha\) decreases without explicit dependency on \texttt{d\_model}.}
\label{tab:gpt-embeddings-norms}
\end{table*}

While Appendix~\ref{sec:appendix} provides estimations for \texttt{ margin} based solely on \(N\), intuitively, the optimal \texttt{margin} should also depend on the dimension size of the embedding space \texttt{d\_model}. To estimate the dependence of \texttt{margin} on \texttt{d\_model}, we propose the following idea.

The intuition is that the effective \texttt{margin} should prevent embedding \(w_i\) from being marginalized in as many non-relevant contexts as possible. To model this behavior of the effective \texttt{margin}, let us denote the effective margin for each \(w_i\) as \(\texttt{margin}_i\) and assume that we want \(\texttt{margin}_i\) to prevent \(w_i\) from marginalization in 95\% of non-relevant contexts. In other words:
\[
\langle h_t, w_i \rangle < \langle h_t, w_{x_t} \rangle - \texttt{margin}_i
\]

should be true for 95\% of \((h_t, w_{x_t})\). We can rewrite this condition as:
\[
\texttt{margin}_i < \langle h_t, w_{x_t} - w_i \rangle 
\]
Let \(\mathcal{P}_{0.05}\) denote the 5th percentile operator. Then we want:
\[
\texttt{margin}_i = \mathcal{P}_{0.05}(\langle h_t, w_{x_t} - w_i \rangle)
\]

Having obtained \(\texttt{margin}_i\), we can estimate the average effective \(\overline{\texttt{margin}}\) as the average of all \(\texttt{margin}_i\):
\[
    \overline{\texttt{margin}} = \frac{1}{N} \sum_{w_i \in W}  \texttt{margin}_i
\]

To sample \((h_t, w_{x_t})\) for each possible \(w_{x_t}\) we take \(h_t = \texttt{LayerNorm}(\gamma \circ w_t)\), where \texttt{LayerNorm} is the final layer normalization in the transformer and \(\gamma\) is the weight in this layer normalization. This \(h_t\) gives \(\langle h_t, w_t \rangle = \max_{h}(\langle h, w_t \rangle) \) \cite{brody-etal-2023-expressivity}.

In Table~\ref{tab:gpt-embeddings-norms}, we show an estimate of the effective \texttt{margin} for pre-trained GPT-2 models and for randomly initialized versions of the model. 

We observe that, with random initialization, the effective \texttt{margin} grows approximately as \(0.02\times\texttt{d\_model}\). However, for trained models, the dependence is more complex, but still increases with increasing \texttt{d\_model}. This suggests that as \texttt{d\_model} increases, the same fixed \texttt{margin} (e.g., the \texttt{margin} from Appendix~\ref{sec:appendix}) will become effective and provide more positive effects.

We noticed that the margin score mainly depends on \(\lVert h_t \rVert \lVert w_{x_t}\rVert\). Therefore, we tried to use:
\[
\overline{\alpha} = \frac{1}{N} \sum_{w_i \in W} \mathcal{P}_{0.05}\left(\frac{\langle h_t, w_{x_t} - w_i \rangle}{\lVert h_t \rVert \lVert w_{x_t}\rVert}\right)
\]
This estimation of \(\alpha\) remains almost constant with changes in \texttt{ d\_model} for a randomly initialized model. The metrics for experiments using the hyperparameter \(\alpha\) can be found in Table~\ref{tab:modifications-quality}.

\section{PyTorch implementations}
\label{sec:appendix-torch}

\begin{lstlisting}[language=Python, caption=PyTorch implementation of Proposed method]
import torch

def thresholding(logits, targets, margin):
    threshold = torch.gather(logits, 2, targets.unsqueeze(-1)) - margin
    logits = torch.where(logits < threshold, torch.tensor(-float('Inf'), device=logits.device), logits)
    return logits
\end{lstlisting}

\begin{lstlisting}[language=Python, caption=PyTorch implementation of Separated Embeddings]
import torch
import torch.nn as nn

class SeparatedEmbedding(nn.Module):
    def __init__(self, num_embeddings, embedding_dim):
        super().__init__()
        self.weights = nn.ParameterList([
            nn.Parameter(
                torch.randn(embedding_dim)) for _ in range(num_embeddings)
        ])

    def forward(self, input):
        weight = torch.stack([w for w in self.weights]).requires_grad_(True)
        return weight[input]
\end{lstlisting}

\section{Hyperparameters}
\begin{table}
  \centering
  \begin{tabular}{|l|c|}
    \hline
    \textbf{Hyperparamenter} & \textbf{Value} \\
    \hline
    block\_size & 64 \\
    batch\_size & 12 \\
    n\_layer & 4 \\
    n\_head & 4 \\
    n\_embd & 128 \\
    max\_iters & 8000 \\
    lr\_decay\_iters & 8000 \\
    dropout & 0 \\
    eval\_iters & 20 \\
    eval\_interval & 250 \\
    learning\_rate & 1e-3 \\
    min\_lr & 1e-4 \\
    weight\_decay & 1e-1 \\
    beta1 & 0.9 \\
    beta2 & 0.99 \\
    grad\_clip & 1.0 \\
    \hline
  \end{tabular}
  \caption{Hyperparameters for the GPT-2 arcitecture model used for experiments.}
  \label{tab:hyperparameters}
\end{table}

The hyperparameters used for the experiments are listed in Table~\ref{tab:hyperparameters}.

\section{PPL vs. Temperature}
\label{sec:ppl-temperature}
\begin{figure}[t]
  \includegraphics[width=\columnwidth]{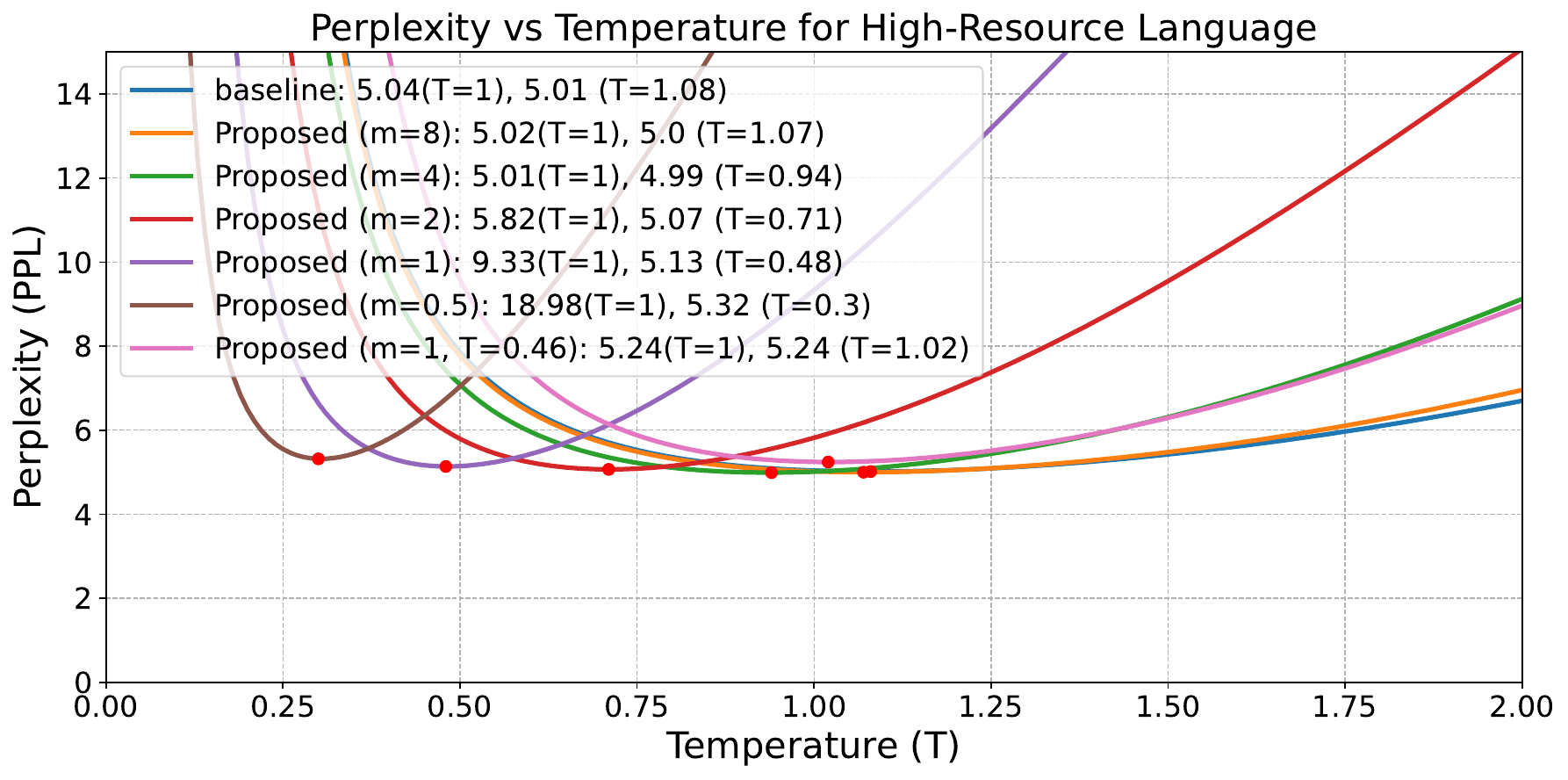}
  \caption{\textbf{PPL} vs. Temperature for high-resource language. The plot shows the \textbf{PPL} of different models as a function of temperature (T) for a high-resource language, calculated for $P_{\theta, T}$. \textbf{PPL} values are plotted for each model, highlighting the minimum perplexity achieved at various temperature levels.}
  \label{fig:ppl-by-t-big}
\end{figure}

\begin{figure}[t]
  \includegraphics[width=\columnwidth]{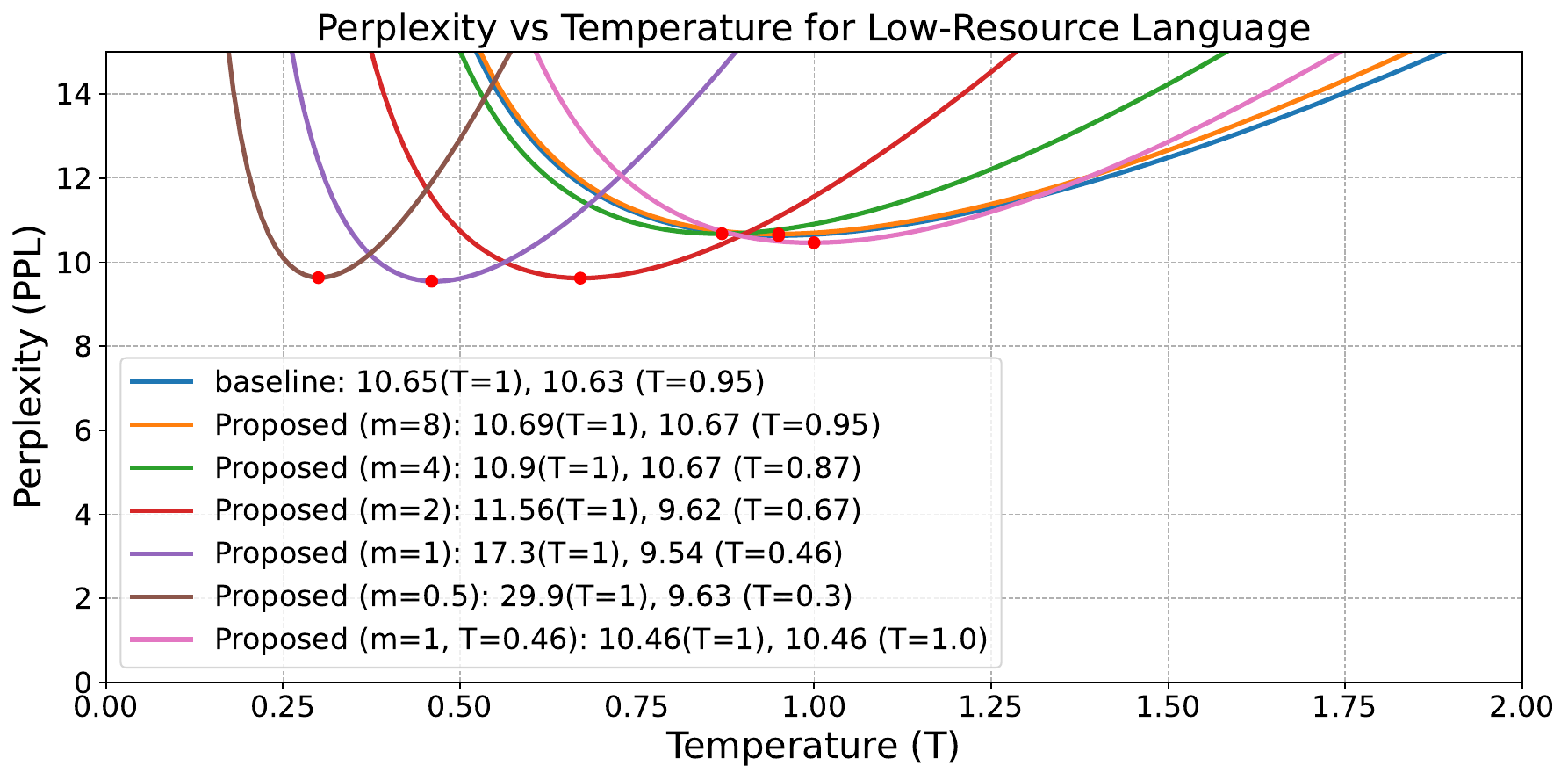}
  \caption{\textbf{PPL} vs. Temperature for low-resource language. The plot shows the \textbf{PPL} of different models as a function of temperature (T) for a low-resource language, calculated for $P_{\theta, T}$. \textbf{PPL} values are plotted for each model, highlighting the minimum perplexity achieved at various temperature levels.}
  \label{fig:ppl-by-t-small}
\end{figure}

Figures~\ref{fig:ppl-by-t-big} and~\ref{fig:ppl-by-t-small} illustrate the relationship between \textbf{PPL} and \(T\) for low- and high-resource languages. The plots highlight how the perplexity of different models, calculated for $P_{\theta, T}$, changes as the temperature is adjusted. These visualizations emphasize the impact of temperature scaling on model performance, with each model achieving its optimal \textbf{PPL} at different temperature values.

The issue can be addressed with some decrease in quality by optimizing $P'_{\theta} = P_{\theta, 1/T_{best}}$. The experiment shows that $T'_{best}$ is 1 for $P'_{\theta}$.

\section{Weight tying}
\label{sec:weight-tying}

Although we justified the usage of the proposed method using weight tying, the proposed method can also be applied to models without weight tying. Table~\ref{tab:wo-weight-tying-quality} shows that the proposed method slightly improves the metrics for models without weight tying.

\section{Other Modifications}

\subsection{Softminus}

We tried to subtract $e^{-\text{margin}}$ from all $P_{\theta}(v_i)$ values above the margin:
\[
P'_{\theta}(v_i) = \max(0, P_{\theta}(v_i) - e^{-\text{margin}}).
\]
This operation ensures that $P'_{\theta}(v_i)~\to~0$ when $P_{\theta}(v_i)~\to~e^{-\text{margin}}$. The results of this experiment are presented in Table~\ref{tab:softminus-results}.

\subsection{Detached Logits Under Threshold}

We attempted to eliminate logits below the threshold, but instead of fully removing them, we detached them to prevent marginalization gradient flow for rare tokens. The results of this experiment are shown in Table~\ref{tab:DUT-results}.

\begin{table*}[!ht]
\centering
  \begin{tabular}{|l|c|c|c|c|c|c|c|}
    \hline
    \textbf{Model} & \textbf{Lang} & \textbf{PPL} & $\textbf{PPL}_{best}(\text{T}_{best})$ & \textbf{Accuracy} & \textbf{Recall@5} & \textbf{MRR} & \textbf{I(W)} \\
    \hline
    \textbf{Baseline} & HR & 5.04 & 5.01 (1.08) & 0.5187 & 0.8299 & 0.6541 & 0.8422 \\
    & LR & 10.65 & 10.63 (0.95)  & 0.3147 & 0.6883 & 0.4803 & 0.4173 \\
    \hline
    \textbf{AGG} & HR & 5.01 & 4.99 (1.09) & 0.5181 & 0.8303 & 0.6540 & 0.8142 \\
    (\(\alpha=0.02\), K=1600) & LR & 11.82 & 11.81 (0.97) & 0.2915 & 0.6691 & 0.4605 & 0.5298 \\
    \hline
    \textbf{AGG} & HR & 5.01 & 4.98 (1.08) & 0.5212 & 0.8308 & 0.6560 & 0.7926 \\
    (\(\alpha=0.2\), K=1600) & LR & 12.24 & 12.21 (0.95) & 0.2875 & 0.6539 & 0.4539 & 0.2600 \\
    \hline
    \textbf{CosReg} & HR & 5.01 & 4.99 (1.08) & 0.5199 & 0.8329 & 0.6558 & 0.8381 \\
    (\(\gamma=1\)) & LR & 10.64 & 10.62 (0.95) & 0.3177 & 0.6849 & 0.4817 & 0.5450 \\
    \hline
    \textbf{Adv} & HR & 5.04 & 5.02 (1.08) & 0.5204 & 0.8294 & 0.6551 & 0.8275 \\
    (\(\alpha=0.05\)) & LR & 10.81 & 10.77 (0.94) & 0.3137 & 0.6857 & 0.4789 & 0.2514 \\
    \hline
    \textbf{CWT} & HR & 7.45 & 7.44 (1.04) & 0.4133 & 0.7333 & 0.5586 & 0.7710 \\
     & LR & 17.70 & 17.67 (0.95) & 0.1949 & 0.4896 & 0.3442 & 0.6417 \\
    \hline
    \textbf{Proposed} & HR & 9.33 & 5.13 (0.48) & 0.5297 & 0.8344 & 0.6626 & 0.8884\\
    (\texttt{margin}=1) & LR & 17.30  & 9.54 (0.46) & 0.3714 & 0.7204 & 0.5255 & 0.7651 \\
    \hline
    \textbf{Proposed} & HR & 18.98 & 5.32 (0.30) & 0.5208 & 0.8315 & 0.6554 & 0.8916 \\
    (\texttt{margin}=0.5) & LR & 29.90 & 9.63 (0.30) & 0.3716 & 0.7229 & 0.5269 & 0.7767 \\
    \hline
    \textbf{Proposed} & HR & 109.71 & 18.62 (0.02) & 0.1610 & 0.5862 & 0.3574 & \textbf{0.9581} \\
    (\texttt{margin}=0) & LR & 100.66 & 43.92 (0.12) & 0.0594 & 0.4065 & 0.2162 & \textbf{0.9352} \\
    \hline
    \textbf{Proposed} & HR & 5.24 & 5.24 (\textbf{1.02}) & 0.5241 & 0.8323 & 0.6579 & 0.8207 \\
    (\texttt{margin}=1, T=0.46) & LR & 10.46  & 10.46 (\textbf{1.00}) & 0.3459 & 0.7064 & 0.5060 & 0.6730 \\
    \hline
    \textbf{Proposed+SE} & HR & 9.18 & 5.12 (0.48) & 0.5274 & \textbf{0.8357} & 0.6615 & 0.8706 \\
    (\texttt{margin}=1) & LR & 13.91  & 6.90 (0.44) & 0.4544 & 0.7827 & 0.5986 & 0.7109 \\
    \hline
    \textbf{Proposed+SE} & HR & 5.22 & 5.19 (1.07) & 0.5243 & 0.8326 & 0.6584 & 0.8552 \\
    (\texttt{margin}=1, T=0.44) & LR & 10.61  & 10.58 (1.05) & 0.3439 & 0.7044 & 0.5041 & 0.5944 \\
    \hline
    \textbf{Proposed+\(\alpha\)} & HR & 5.36 & \textbf{4.97} (0.77) & \textbf{0.5317} & 0.8354 & \textbf{0.6640} & 0.8660 \\
    (\(\alpha\)=0.25) & LR & 10.61  & 9.25 (0.71) & 0.3720 & 0.7239 & 0.5266 & 0.6823 \\
    \hline
    \textbf{Proposed+\(\alpha\)} & HR & 32.70 & 5.29 (0.19) & 0.5213 & 0.8333 & 0.6572 & 0.9061 \\
    (\(\alpha\)=0.0625) & LR & 50.38  & 9.46 (0.18) & 0.3759 & 0.7244 & 0.5296 & 0.7812 \\
    \hline
    \textbf{Proposed+\(\alpha\)+SE} & HR & 31.75 & 5.25 (0.19) & 0.5243 & 0.8330 & 0.6588 & 0.8895 \\
    (\(\alpha\)=0.0625) & LR & 38.50  & \textbf{6.11} (0.17) & \textbf{0.4885} & \textbf{0.8132} & \textbf{0.6299} & 0.8298 \\
    \hline
  \end{tabular}
  \caption{Extended table with results from other papers and the proposed method with different hyperparameters.}
  \label{tab:modifications-quality}
\end{table*}

\begin{table*}[!ht]
\centering
  \begin{tabular}{|l|c|c|c|c|c|c|c|}
    \hline
    \textbf{Model} & \textbf{Lang} & \textbf{PPL} & $\textbf{PPL}_{best}(\text{T}_{best})$ & \textbf{Accuracy} & \textbf{Recall@5} & \textbf{MRR} & \textbf{I(W)} \\
    \hline
    \textbf{Baseline} & HR & \textbf{4.98} & \textbf{4.94} (1.11) & 0.5240 & 0.8329 & 0.6583 & 0.8157 \\
    (w/o WT) & LR & 8.82 & 8.79 (0.94)  & 0.3725 & 0.7247 & 0.5285 & 0.5403 \\
    \hline
    \textbf{Proposed} & HR & 5.80 & 5.07 (0.71) & 0.5268 & 0.8333 & 0.6603 & 0.8699 \\
    (w/o WT, \texttt{margin}=2) & LR & 10.75  & 8.94 (0.67) & \textbf{0.3784} & \textbf{0.7297} & \textbf{0.5342} & 0.6188 \\
    \hline
    \textbf{Proposed} & HR & 9.43 & 5.12 (0.48) & 0.5292 & \textbf{0.8364} & 0.6623 & 0.8693 \\
    (w/o WT, \texttt{margin}=1) & LR & 16.43  & 9.14 (0.47) & 0.3781 & 0.7285 & 0.5331 & \textbf{0.6985} \\
    \hline
    \textbf{Baseline+SE} & HR & 4.99 & 4.95 (1.11) & 0.5219 & 0.8316 & 0.6567 & 0.6395 \\
    (w/o WT) & LR & \textbf{8.71} & \textbf{8.71} (0.98)  & 0.3722 & 0.7281 & 0.5279 & 0.3457 \\
    \hline
    \textbf{Proposed+SE} & HR & 5.61 & 4.98 (0.72) & \textbf{0.5304} & 0.8344 & \textbf{0.6630} & 0.7570 \\
    (w/o WT, \texttt{margin}=2) & LR & 10.42  & 9.00 (0.7) & 0.3700 & 0.7294 & 0.5283 & 0.6182 \\
    \hline
    \textbf{Proposed+SE} & HR & 8.83 & 5.07 (0.49) & 0.5284 & 0.8352 & 0.6623 & \textbf{0.8721} \\
    (w/o WT, \texttt{margin}=1) & LR & 15.37  & 9.11 (0.49) & 0.3767 & 0.7289 & 0.5321 & 0.6732 \\
    \hline
  \end{tabular}
  \caption{Results for models without weight tying.}
  \label{tab:wo-weight-tying-quality}
\end{table*}

\begin{table*}[!ht]
\centering
  \begin{tabular}{|l|c|c|c|c|c|c|c|}
    \hline
    \textbf{Model} & \textbf{Lang} & \textbf{PPL} & $\textbf{PPL}_{best}(\text{T}_{best})$ & \textbf{Accuracy} & \textbf{Recall@5} & \textbf{MRR} & \textbf{I(W)} \\
    \hline
    \textbf{Proposed} & HR & 5.82 & 5.07 (0.71) & 0.5291 & 0.8336 & 0.6614 & 0.8499\\
    (\texttt{margin}=2) & LR & 11.56 & 9.62 (0.67)  & 0.3581 & 0.7166 & 0.5161 & 0.6422 \\
    \hline
    \textbf{Proposed} & HR & 9.33 & 5.13 (0.48) & \textbf{0.5297} & 0.8344 & \textbf{0.6626} & 0.8884\\
    (\texttt{margin}=1) & LR & 17.30  & 9.54 (0.46) & 0.3714 & 0.7204 & 0.5255 & \textbf{0.7651} \\
    \hline
    \textbf{Softminus} & HR & \textbf{5.81} & 5.15 (0.73) & 0.5287 & 0.8327 & 0.6604 & 0.8733 \\
    (\texttt{margin}=2) & LR & \textbf{11.47}  & 9.91 (0.71) & 0.3623 & 0.7086 & 0.5169 & 0.6384 \\
    \hline
    \textbf{Softminus} & HR & 9.06 & 5.30 (0.51) & 0.5281 & 0.8323 & 0.6604 & 0.8774 \\
    (\texttt{margin}=1) & LR & 16.38  & 9.74 (0.5) & 0.3750 & 0.7208 & 0.5279 & 0.7488 \\
    \hline
    \textbf{Softminus} & HR & 14.21 & \textbf{4.95} (0.35) & \textbf{0.5297} & \textbf{0.8352} & 0.6623 & \textbf{0.9048} \\
    (\texttt{margin}=2, C=5) & LR & 25.44  & 9.13 (0.32) & 0.3773 & 0.7233 & 0.5299 & 0.7358 \\
    \hline
    \textbf{Softminus} & HR & 30.72 & 4.99 (0.20) & 0.5283 & 0.8336 & 0.6615 & 0.8844 \\
    (\texttt{margin}=1, C=5) & LR & 46.76  & \textbf{9.04} (0.18) & \textbf{0.3829} & \textbf{0.7249} & \textbf{0.5339} & 0.6975 \\
    \hline
    \textbf{Softminus} & HR & 20.96 & 10.83 (0.47) & 0.3741 & 0.6976 & 0.5218 & 0.0000 \\
    (DT, \texttt{margin}=2) & LR & 49.71  & 39.13 (0.55) & 0.1490 & 0.3650 & 0.2726 & 0.0000 \\
    \hline
    \textbf{Softminus} & HR & 47.19 & 18.71 (0.31) & 0.2415 & 0.5607 & 0.3858 & 0.0000 \\
    (DT, \texttt{margin}=1) & LR & 97.10  & 91.01 (0.59) & 0.0000 & 0.1434 & 0.0803 & 0.0000 \\
    \hline
    \textbf{Softminus} & HR & 17.50 & 10.95 (0.39) & 0.3228 & 0.6882 & 0.5017 & 0.4870 \\
    (DT, \texttt{margin}=2, C=5) & LR & 44.69  & 31.67 (0.42) & 0.1242 & 0.3982 & 0.2620 & 0.1795 \\
    \hline
    \textbf{Softminus} & HR & 33.91 & 10.55 (0.2) & 0.3463 & 0.6956 & 0.6595 & 0.5017 \\
    (DT, \texttt{margin}=1, C=5) & LR & 81.32  & 49.20 (0.22) & 0.1417 & 0.3162 & 0.2464 & 0.2880 \\
    \hline
  \end{tabular}
  \caption{Results for the softminus method. DT stants for detached threshold, an experiment where the threshold was detached before subtracting it from logits.}
  \label{tab:softminus-results}
\end{table*}

\begin{table*}[!ht]
\centering
  \begin{tabular}{|l|c|c|c|c|c|c|c|}
    \hline
    \textbf{Model} & \textbf{Lang} & \textbf{PPL} & $\textbf{PPL}_{best}(\text{T}_{best})$ & \textbf{Accuracy} & \textbf{Recall@5} & \textbf{MRR} & \textbf{I(W)} \\
    \hline
    \textbf{Proposed} & HR & inf & inf (1.00) & 0.4148 & 0.7565 & 0.5638 & 0.0002 \\
    (DUT, \texttt{margin}=2) & LR & 18.39 & 18.25 (1.11) & 0.2343 & 0.5446 & 0.3810 & 0.2724 \\
    \hline
    \textbf{Proposed} & HR & 15.78 & 15.71 (0.92) & 0.2757 & 0.6368 & 0.4386 & 0.7866 \\
    (DUT, \texttt{margin}=1) & LR & 29.09  & 28.98 (0.92) & 0.1491 & 0.3771 & 0.2819 & 0.3075 \\
    \hline
  \end{tabular}
  \caption{Results for experiments where logits under the threshold were not eliminated but only detached. DUT stands for detached under threshold.}
  \label{tab:DUT-results}
\end{table*}
\end{document}